\documentclass[sigconf]{aamas} 

\usepackage{balance} 
\usepackage[ruled,vlined]{algorithm2e}
\usepackage{multirow}

\DeclareMathOperator*{\argmin}{arg\,min}
\DeclareMathOperator*{\argmax}{arg\,max}

\setcopyright{none}
\renewcommand\footnotetextcopyrightpermission[1]{} 
\setcopyright{ifaamas}
\acmConference[AAMAS '22]{Proc.\@ of the 21st International Conference
on Autonomous Agents and Multiagent Systems (AAMAS 2022)}{May 9--13, 2022}
{Auckland, New Zealand}{P.~Faliszewski, V.~Mascardi, C.~Pelachaud,
M.E.~Taylor (eds.)}
\copyrightyear{2022}
\acmYear{2022}
\acmDOI{}
\acmPrice{}
\acmISBN{}

\acmSubmissionID{349}

\title[Multi-Objective Reinforced Active Learning]{MORAL: Aligning AI with Human Norms through Multi-Objective Reinforced Active Learning}

\author{Markus Peschl}
\affiliation{
  \institution{Delft University of Technology}
  \city{Delft}
  \country{The Netherlands}}
\email{peschl@protonmail.com}

\author{Arkady Zgonnikov}
\affiliation{
  \institution{Delft University of Technology}
  \city{Delft}
  \country{The Netherlands}}
\email{A.Zgonnikov@tudelft.nl}

\author{Frans A. Oliehoek}
\affiliation{
  \institution{Delft University of Technology}
  \city{Delft}
  \country{The Netherlands}}
\email{F.A.Oliehoek@tudelft.nl}

\author{Luciano C. Siebert}
\affiliation{
  \institution{Delft University of Technology}
  \city{Delft}
  \country{The Netherlands}}
\email{L.CavalcanteSiebert@tudelft.nl}

\begin{abstract}
Inferring reward functions from demonstrations and pairwise preferences are auspicious approaches for aligning Reinforcement Learning (RL) agents with human intentions. However, state-of-the art methods typically focus on learning a single reward model, thus rendering it difficult to trade off different reward functions from multiple experts. We propose Multi-Objective Reinforced Active Learning (MORAL), a novel method for combining diverse demonstrations of social norms into a Pareto-optimal policy. Through maintaining a distribution over scalarization weights, our approach is able to interactively tune a deep RL agent towards a variety of preferences, while eliminating the need for computing multiple policies. We empirically demonstrate the effectiveness of MORAL in two scenarios, which model a delivery and an emergency task that require an agent to act in the presence of normative conflicts. Overall, we consider our research a step towards multi-objective RL with learned rewards, bridging the gap between current reward learning and machine ethics literature.
\end{abstract}


\newcommand{\BibTeX}{\rm B\kern-.05em{\sc i\kern-.025em b}\kern-.08em\TeX}

\begin{document}


\pagestyle{fancy}
\fancyhead{}


\maketitle

\section{Introduction}
\noindent The design of adequate reward functions poses a tremendous challenge for building reinforcement learning (RL) agents that ought to act in accordance with human intentions \cite{Amodei2016, everitt2017reinforcement}. Besides complicating the deployment of RL in the real world \cite{dulac2019challenges}, this can lead to major unforeseen societal impacts, which need to be accounted for when building autonomous systems \cite{Whittlestone2021, nickbostrom2014}. To tackle this, the field of value alignment has largely focused on reward learning, which aims to adopt a bottom-up approach of finding goal specifications from observational data instead of manually specifying them \cite{Hadfield-Menell2016, noothigattu2019teaching, shah2019feasibility}. However, such technical approaches cannot on their own solve the normative value alignment problem of deciding which values should ultimately be encoded into an agent \cite{gabriel2020artificial}. Nonetheless, building methods that allow for learning and trading off different conflicting values could potentially alleviate this issue, thus making such methods an important avenue of research for beneficial artificial intelligence (AI) \cite{russell2015research}.

Jointly optimizing for different rewards can be cast into multi-objective RL (MORL) \cite{roijers2013}, which constitutes a promising framework for building human-aligned AI \cite{vamplew2018human}. Using game-theoretic notions of optimality, MORL typically aims to find a solution, or a set thereof, that can represent a variety of preferences over the components of a vector-valued reward function. While this can theoretically tackle the overoptimization of narrowly defined tasks, the designer still needs to manually specify multiple reward functions. 

Inverse RL (IRL) \cite{Ziebart2008, gaurav2019discriminatively} and preference-based RL \cite{Wirth2017, Christiano2017} offer techniques for avoiding the reward design problem altogether by learning a parametric reward model from demonstrations and pairwise preferences, respectively. In this paper, we combine these approaches in a multi-objective setting with a focus on learning human norms. The motivation for this is twofold: Firstly, previous research on learning multiple reward functions has mostly employed latent-variable IRL models for finding multiagent \cite{gruver2020multi}, hierarchical \cite{venuto2020oirl, sharma2018directed} and multitask \cite{yu2019meta, gleave2018multi} rewards, whereas finding aggregated rewards from conflicting sequential data has yet to be addressed. Secondly, a major challenge of value alignment is ensuring that an agent can predict its adherence to norms in the environment. Such adherence is implicitly embedded in human goal specifications but missing in manually engineered reward functions \cite{hadfield2019incomplete}. Our goal therefore is to find a policy that acts on a common set of social norms while allowing for fine-tuning the agent with respect to inherent disagreements that may arise. 


\textbf{Contributions}. We propose Multi-Objective Reinforced Active Learning (MORAL), a method that combines active preference learning and IRL to interactively learn a policy of social norms from expert demonstrations. MORAL first finds a vector-valued reward function through adversarial IRL, which is subsequently used in an interactive MORL loop. By requesting pairwise preferences over trajectories of on-policy experience from an expert, MORAL learns a probability distribution over linear combinations of reward functions under which the optimal policy most closely matches the desired behavior. We show that our approach directly approximates a Pareto-optimal solution in the space of expert reward functions, without the need of enumerating through a multitude of preference weights. Finally, we demonstrate that MORAL efficiently captures norms in two gridworld scenarios, while being able to adapt the agent's behavior to a variety of preferences.\footnote{Source code is available at \url{https://github.com/mlpeschl/moral_rl}.}

\section{Preliminaries}
\noindent \textbf{Multi-Objective RL.}
We employ a multi-objective Markov decision process (MOMDP) for framing the problem of aligning an agent with multiple experts. Formally, a MOMDP is given by the tuple $\langle \mathcal{S}, \mathcal{A}, p, \mathbf{r}, \mu_0, \gamma\rangle$, with state space $\mathcal{S}$, the set of actions $\mathcal{A}$, a transition distribution $p(s'|s,a)$, a vector-valued reward function $\mathbf{r}(s,a) \in \mathbb{R}^m$, a starting state distribution $\mu_0$ and the discount factor $\gamma \in [0,1)$. We consider optimal solutions to be given by a Pareto frontier $\mathcal{F} = \{\pi | \nexists \pi' \neq \pi  : J_\mathbf{r}(\pi') \geq J_\mathbf{r}(\pi) \}$, where $J_\mathbf{r}(\pi) = \mathbb{E}_\pi[ \sum_{t=0}^T \gamma^t \mathbf{r}(s_t,a_t)]$ is the vector-valued return of a policy $\pi: \mathcal{S} \rightarrow \Delta_{\mathcal{A}}$ that maps states to a distribution over actions. Furthermore, we define the convex coverage set (CCS) $\mathcal{F}^* = \{\pi \in \mathcal{F} \ \vert\ \exists\mathbf{w}\in \mathbb{R}^m : \mathbf{w}^T J_\mathbf{r}(\pi) \geq\mathbf{w}^T J_\mathbf{r}(\pi'), \ \forall \pi'\in \mathcal{F}\}$ to be the subset of Pareto-optimal solutions that can be obtained through optimizing for linear combinations of the different reward components. 
\vspace{4pt}

\noindent \textbf{Proximal Policy Optimization (PPO).}
Given a weight $\mathbf{w}\in \mathbb{R}^m$, we can optimize for policies on the CCS using PPO \cite{schulman2017proximal} on the scalarized reward $r(s,a) = \mathbf{w}^T \mathbf{r}(s,a)$. Using on-policy experience, PPO maximizes the return of a parametrized policy $\pi_\phi$ by performing gradient descent on the clipped objective 
$$
    \mathcal{L}^{\text{CLIP}}(\phi) = \mathbb{E}_t[\min(r_t(\phi)\hat{A}_t, clip(r_t(\phi), 1-\epsilon, 1+\epsilon)\hat{A}_t)],
$$
where $\mathbb{E}_t$ is the expectation at time $t$, $r_t$ is a ratio of the new versus the current policy, $\hat{A}_t$ is an estimated advantage at time $t$ and $clip(x,a,b)$ limits the value of $x$ to the interval $[a,b]$.

\vspace{4pt}

\noindent \textbf{Adversarial IRL (AIRL).} The maximum entropy IRL \cite{Ziebart2008} goal is to derive a reward function $r_\theta$ from a demonstration dataset $\mathcal{D} = \{\tau_i\}_{i=1}^N$ of expert  trajectories $\tau = \{s_t, a_t\}_{t=0}^T$ by solving a maximum likelihood problem 
$\max_\phi \mathbb{E}_{\tau\sim \mathcal{D}}[\log p_\theta(\tau)]$, where
\begin{equation}
p_\theta(\tau) \propto \mu_0(s_0) \prod_{t=0}^T p(s_{t+1} | s_t, a_t)\exp({r_\theta(s_t, a_t)}) = \overline{p}_\theta(\tau). \label{eq:maxentdistrib}
\end{equation}
AIRL \cite{Fu2017} approximately solves the IRL problem using generative adversarial networks \cite{Goodfellow2014GenerativeAN}. It jointly trains a policy (generator) $\pi_\phi$ alongside a discriminator of the form 
$$
    D_{\theta}(s,a) = \frac{\exp(f_\theta(s,a))}{\exp(f_\theta(s,a)) + \pi_\phi(a|s)}.
$$
While $D_\theta$ is trained using a binary cross-entropy loss to distinguish trajectories in $\mathcal{D}$ from $\pi_\phi$, the agent maximizes its returns using the reward $r(s,a) = \log (D_\theta(s,a)) -\log (1-D_\theta(s,a))$. 

\section{Multi-Objective Reinforced Active Learning}

\begin{figure}[t!]
    \centering
    \includegraphics[width=0.99\columnwidth]{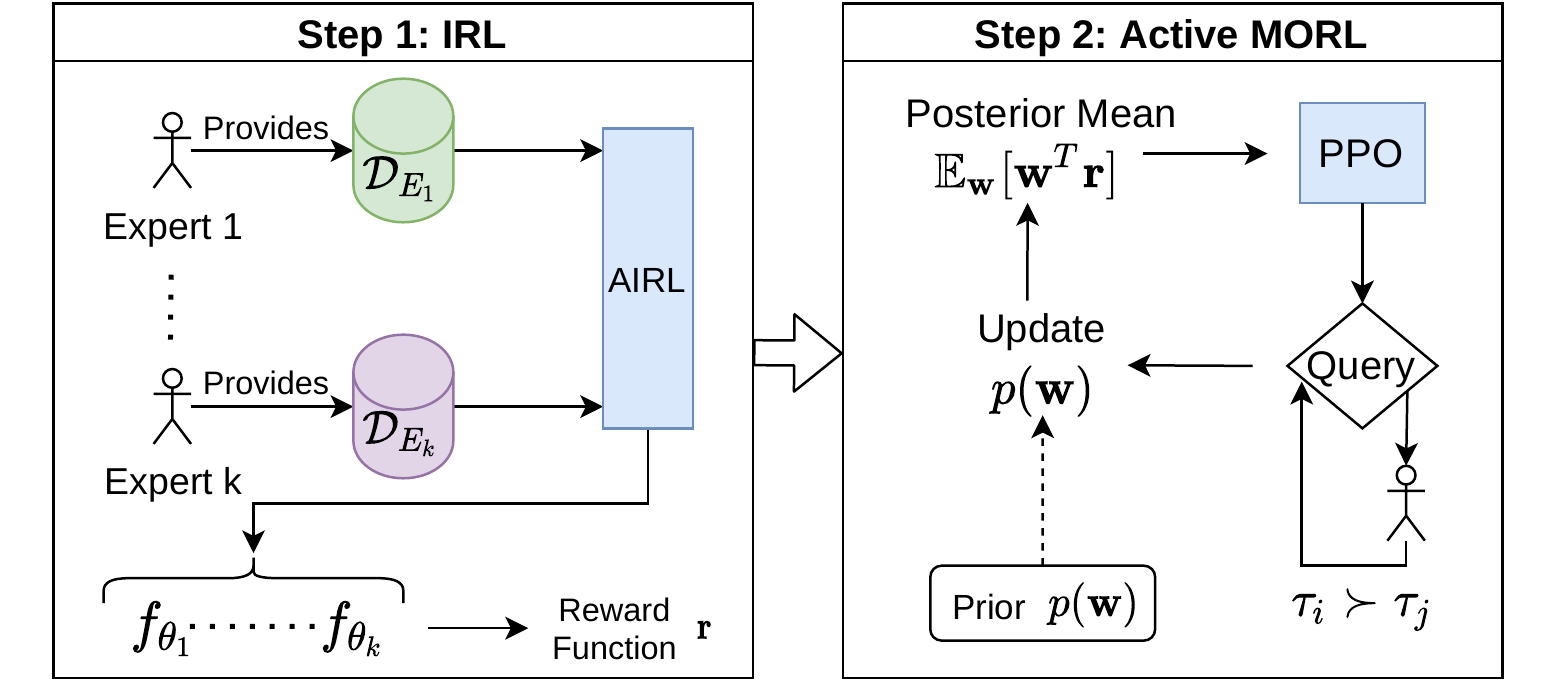}
    \caption{Multi-Objective Reinforced Active Learning.}
    \label{fig:moral_illustration}
\end{figure}

MORAL uses a two-step procedure that separates reward and policy training to learn from multiple experts (figure~\ref{fig:moral_illustration}). In the first step (\textit{IRL}), we use a set of trajectories $\mathcal{D}_E = \cup_{i=1}^k \mathcal{D}_{E_i}$ demonstrated by $k$ distinct experts and run AIRL to obtain a vector of reward functions $\mathbf{r} = (f_{\theta_1}, \dots, f_{\theta_k})$ and imitation policies $(\pi^*_{E_1}, \dots, \pi^*_{E_k})$ for each subset of trajectories $\mathcal{D}_{E_i}$. In step two (\textit{active MORL}), we run an interactive MORL algorithm for learning a distribution over weights that determine a linear combination of the different components in $\mathbf{r}$. We will now explain how this distribution is learned alongside training a deep RL agent in a single loop. Firstly, active MORL takes a prior $p(\mathbf{w})$ over scalarization weights and initializes a reward function $r(s,a)= \mathbb{E}_\mathbf{w}[\mathbf{w}^T\mathbf{r}(s,a)]$. Subsequently, we repeatedly (i) \textit{optimize} for the scalar reward $r$ by running PPO for a given number of steps, (ii) \textit{query} an expert for a pairwise comparison $q_n$ of two trajectories and (iii) \textit{update} the posterior $p(\mathbf{w} | q_1, \dots, q_n)$. Finally, the reward function is reinitialized to the posterior mean scalarization and is used by PPO in the next iteration.

To \textit{query} and \textit{update}, we specify a probabilistic model over expert preferences. We employ a Bradley-Terry model \cite{bradley1952rank}
\begin{equation}
    p(\tau_i \succ \tau_j|\mathbf{w}) = \frac{\exp({\mathbf{w}^T \mathbf{r}(\tau_i)})}{\exp({\mathbf{w}^T \mathbf{r}(\tau_j)}) + \exp({\mathbf{w}^T \mathbf{r}(\tau_i)})}, \label{eq:bradley_likelihood}
\end{equation}
with  $\mathbf{r}(\tau) = \sum_{(s, a, s') \in \tau}\mathbf{r}(s, a, s')$ being the reward obtained from a trajectory $\tau$ and $\tau_i\succ \tau_j$ denoting the preference of $\tau_i$ over $\tau_j$. This way, trajectories that achieve a higher linearly scalarized reward are ranked exponentially better in proportion. Assuming that a number of pairwise comparisons $\{q_1 , \dots q_n\}$ have been obtained, we can then simply update the posterior in a Bayesian manner
\begin{equation}
    p(\mathbf{w} | q_1, \dots, q_n) \propto p(\mathbf{w}) \prod_{t=1}^n p(q_t | \mathbf{w}).
    \label{eq:bradley_posterior}
\end{equation}
In our experiments, we choose the prior $p(\mathbf{w})$ to be uniform over all weights $\mathbf{w}$ with $||\mathbf{w}|| \leq 1$ and $\mathbf{w}\geq 0$.

This Bayesian model allows us to maintain a posterior that determines which reward components should be prioritized at each iteration. By providing pairwise preferences, an expert is then able to fine-tune the agent to any specific behavior that can theoretically result from the linear combination of reward components. Furthermore, we can leverage the uncertainty of the posterior to enable the agent to form queries the answers to which are highly informative. Namely, we use the active learning procedure which forms queries based on the amount of removed posterior volume~\cite{Sadigh2017ActivePL}
\begin{equation}
    \max_{(\tau_i,\tau_j)} \min \Big(\mathbb{E}_{\mathbf{w}}[1-p(\tau_i\succ \tau_j| \mathbf{w})], \mathbb{E}_{\mathbf{w}}[1-p(\tau_j\succ \tau_i |\mathbf{w})]\Big), \label{eq:volume_removal}
\end{equation}
where the expectation over $\mathbf{w}$ is approximated using Markov Chain Monte Carlo (MCMC) \cite{chib1995understanding}.

Unfortunately, for sufficiently complex MOMDPs, maximizing this expression over all pairs of feasible trajectories proves to be computationally intractable. Instead, we do a discrete search over randomly sampled pairs of trajectories that arise during on-policy RL experience (algorithm~\ref{alg:moral}). Before each policy improvement step, we sample pairs $(\tau_i, \tau_j)$ and evaluate the corresponding minimum in expression~\eqref{eq:volume_removal}. If $(\tau_i, \tau_j)$ scores highest among all previous pairs obtained since the last posterior update, it is saved in a buffer and queued for the next query, unless a better pair is found later on.

Overall, this active learning scheme minimizes the number of queries needed during training to converge to a desired reward scalarization. Nonetheless, forming queries based on on-policy experience can only lead to locally optimal solutions. Therefore, assuming fixed weights $\mathbf{w}$, we optimize for an entropy-regularized objective
\begin{equation}
    \pi^* = \argmax_\pi \mathbb{E}_\pi\left[\sum_{t=0}^T \mathbf{w}^T \mathbf{r}(s_t,a_t) - \log \pi(s_t, a_t) \right]. \label{eq:pistar}
\end{equation}
This way, MORAL can be interpreted as finding an average of Kullback-Leibler (KL) divergences~\cite{kullback1997information} between the policy distribution over trajectories $\pi(\tau) = \mu_0(s_0)\prod_{t=0}^{T-1}p(s_{t+1} \vert s_t, a_t) \pi(a_t\vert s_t)$ and the marginal maximum entropy IRL distributions (\ref{eq:maxentdistrib}).

\begin{theorem}
\label{thm:moral_kla}
Given $\mathbf{w} \in \mathbb{R}^k$ with $\mathbf{w} \geq 0$, $\sum w_i = 1$, we have that 
\begin{equation}
    \pi^* = \argmin_\pi \sum_{i=1}^k w_i D_{KL}(\pi(\tau)) || p_{\theta_i}(\tau)). \label{eq:kla_moral}
\end{equation}
\end{theorem}
\noindent \textit{Proof:} We provide a proof in the appendix.



Theorem \ref{thm:moral_kla} assumes that all components in $\mathbf{r}$ arise from maximum entropy IRL. However, in practical applications one might want to encode additional prior knowledge into the agent's behavior through a manually engineered primary reward function $r_P$. Nonetheless, by applying analogous reasoning, expression~\eqref{eq:pistar} can be interpreted as maximizing the returns $\mathbb{E}_\pi[\sum_{t=0}^T r_P(s_t,a_t)]$ with a KL regularizer in the form of expression \eqref{eq:kla_moral}. Under this interpretation, MORAL interactively finds regularization hyperparameters that determine which expert's behavior should be prioritized at runtime.

\subsection{Reward Normalization}
Finding scalarization weights in the presence of reward functions with highly different scales is a challenging task for many MORL algorithms. MORAL, on the other hand, learns its weights from preferences, thus making it less susceptible to reward functions that are difficult to compare. Nevertheless, the scale of the reward indirectly impacts the sensitivity of the posterior, since the magnitude of the likelihood (\ref{eq:bradley_likelihood}) depends on the scalar products of the form $\mathbf{w}^T \mathbf{r}(\tau)$. Although $\mathbf{w}$ is bounded by the prior, its product with the vector-valued reward $\mathbf{r}(\tau)$ can become arbitrarily large, which introduces a risk of removing significant parts of the posterior support based on a single query. To tackle this, we utilize the policies obtained from AIRL to normalize each reward component by setting
 \begin{equation}
     \overline{f_{\theta_i}}(s,a) = \frac{f_{\theta_i}(s,a)}{|J(\pi^*_{E_i})|},
 \end{equation}
where $J(\pi^*_{E_i}) = \mathbb{E}_{\pi^*_{E_i}}[\sum_{t=0}^{T}\gamma^t f_{\theta_i}(s,a)]$ is the scalar return of $\pi^*_{E_i}$. This does not introduce any computational overhead, and we simply estimate $J(\pi^*_{E_i})$ by taking the average obtained return with respect to $f_{\theta_i}$ after running AIRL.

\begin{algorithm}[t]
\caption{Multi-Objective Reinforced Active Learning}\label{alg:moral}
 
 \textbf{Input}: Expert demonstrations $\mathcal{D}_E = \{\tau_i\}_{i=1}^N$, prior $p(\mathbf{w})$.\\
 \textbf{Initialize}: Reward function $\mathbf{r} = (f_{\theta_1}, \dots, f_{\theta_k})$ by running AIRL on $\mathcal{D}_E$, PPO agent $\pi_\phi$.\\
 \For{$n = 0, 1, 2,\dots$}{
 Approximate $p(\mathbf{w}| q_1, \dots, q_n)$ through MCMC.\\
 Get mean reward function $r \leftarrow \mathbb{E}_\mathbf{w}[\mathbf{w}^T\mathbf{r}]$.\\
 $volume \leftarrow -\infty$\\
 \For{$k= 0, 1, 2, \dots, N$}{
     Sample trajectories $\mathcal{D} = \{\tau_i\}_{i=1}^m$ using $\pi_\phi$.\\
     Update $\phi$ using PPO to maximize \quad\quad$\mathbb{E}_{\pi_\phi}\left[\sum_{t=0}^{T} \gamma^t{r}(s_t,a_t)\right]$.\\
     Sample a pair of trajectories $(\tau_i, \tau_j)$ from $\mathcal{D}$.\\
    $next\_volume \leftarrow \min (\mathbb{E}_{\mathbf{w}}[1-p(\tau_i\succ \tau_j| \mathbf{w})], \mathbb{E}_{\mathbf{w}}[1-p(\tau_j\succ \tau_i |\mathbf{w})])$ .\\
     \If{$ next\_volume > volume$}{$next\_query \leftarrow (\tau_i, \tau_j)$ \\ $volume \leftarrow next\_volume$}
    }
    Query expert using $next\_query$ and save answer $q_n$.
 }
\end{algorithm}

\section{Experiments}
In the following, we will demonstrate the ability of MORAL in simulation studies of two gridworld environments. To enable a qualitative analysis of the method, we assume that in both environments, a ground truth reward function exists, which is used to generate demonstrations and responses to the agent's queries. Furthermore, by following the experimental setup of related research \cite{noothigattu2019teaching, Wu2018}, we consider environments with a primary reward function $r_P$, encoding a generic task that can easily be solved through deep RL. In this case, we can apply MORAL as before, but add $r_P$ as an additional reward component to the AIRL reward functions for the active learning step. To form the gridworld state, we make a binary array $I\in \{0,1\}^{C\times W\times H}$ of width $W$ and height $H$, as well as channels that encode grid occupancy for all $C$ different object types on the grid. Finally, we employ a convolutional neural network architecture for PPO, consisting of two base convolutional layers with kernel size $2$, and $64$ as well as $256$ output channels respectively. Its activations are then fed into two separate convolutional layers with kernel size $2$ and $32$ output channels each, followed by a linear layer for the critic and actor heads. The details of our implementation are provided in the appendix.

\subsection{Emergency}
We start by illustrating how MORAL can be applied to incorporating social norms from a single expert alongside a primary goal. We define the \textit{Emergency} gridworld as follows: An agent, as well as $6$ humans are randomly positioned onto a $6\times 6$ grid. The humans are lost and need to be escorted before a time limit of $T=75$. Furthermore, the bottom right corner contains a goal state (e.g. a fire extinguisher in a burning room), which the agent uses when standing on its cell. At each step, the agent can either move in one of the four directions, interact with an adjacent cell or do nothing.

We define the agent's primary goal $r_P$ to give a reward of $+0.1$ for each time step spent in the fire extinguisher cell. On the other hand, the social norm of helping people is not considered in $r_P$. In order to learn the latter, we find a reward $f_\theta$ by running AIRL on $50$ synthetic demonstrations coming from a PPO agent that maximizes the number of people saved. Subsequently, we form a reward vector $\mathbf{r} = (r_P, f_\theta)$ and run interactive MORL using a total of $25$ queries. Since we would like to incorporate the goal of saving all people into the primary task of extinguishing fire, we provide preferences in the following way: Given two trajectories $(\tau_i, \tau_j)$, we return $i\succ j$ if the number of people saved in $\tau_i$ exceeds that of $\tau_j$. If both trajectories save the same number of people, we opt for the trajectory that spent more time in the extinguisher cell. Finally, queries are spread out evenly over $6\cdot 10^6$ environment steps.

\begin{figure}[t!]
    \centering
    \includegraphics[width=0.9\columnwidth]{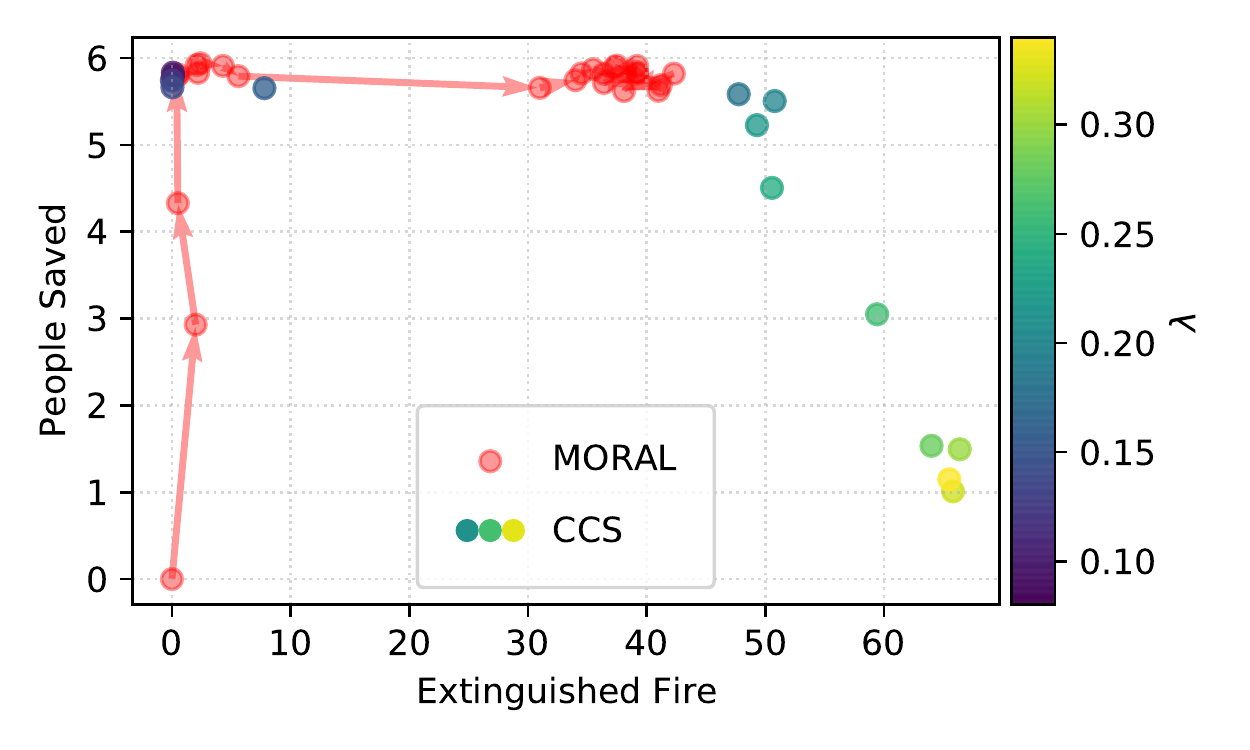}
    \caption{Intermediate policies found during MORAL in the \textit{Emergency} domain, compared to a manually computed CCS. MORAL approximates a Pareto-optimal solution that most closely matches the given preferences.}
    \label{fig:moral_path}
\end{figure}

\begin{figure}[t]
    \centering
    \includegraphics[width=0.75\columnwidth]{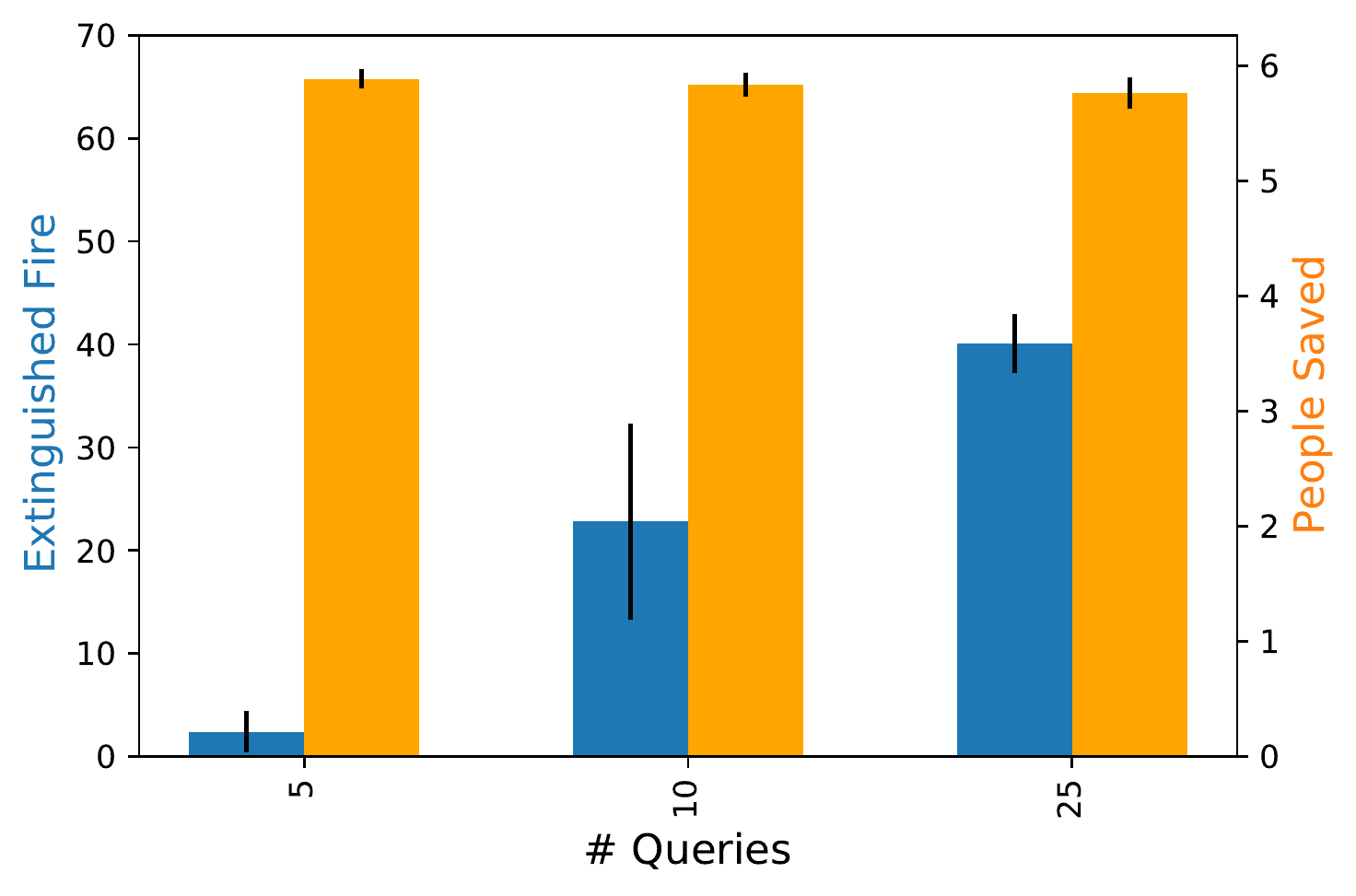}
    \caption{Query efficiency of MORAL for finding a trade-off that matches the given preferences. Averaged over three random seeds.}
    \label{fig:aamas_v2_queries}
\end{figure}

Figure \ref{fig:moral_path} shows the set of policies obtained during training of MORAL and compares it with a CCS found from a manual scalarization $\lambda r_P + (1-\lambda)f_\theta$ for different choices of $\lambda \in [0,1]$. Also, we do not show solutions corresponding to higher values of $\lambda$, since these collapse to a single trivial solution. To illustrate the evolution of solutions, we estimate average returns and plot a corresponding point before each update of the weight posterior. MORAL directly approximates a Pareto-optimal point that opts for saving all people present in the world, while maximizing the agent's performance with respect to the primary goal. Furthermore, MORAL first learns to only save people, which correctly corresponds to the way preferences are provided. Thus, MORAL demonstrates to be successful at directly finding a normative policy while incorporating reward information from multiple sources. To ensure consistency across multiple runs, we also plot the average returns for different numbers of overall queries in figure \ref{fig:aamas_v2_queries}. We see that although $25$ queries are necessary to converge to a solution that closely matches the given preferences, MORAL learns a reasonable trade-off after $10$ queries, which consistently saves all people at the cost of spending less time on the primary goal.

\subsection{Delivery}

\begin{figure}[b!]
    \centering
    \includegraphics[width=0.8\columnwidth]{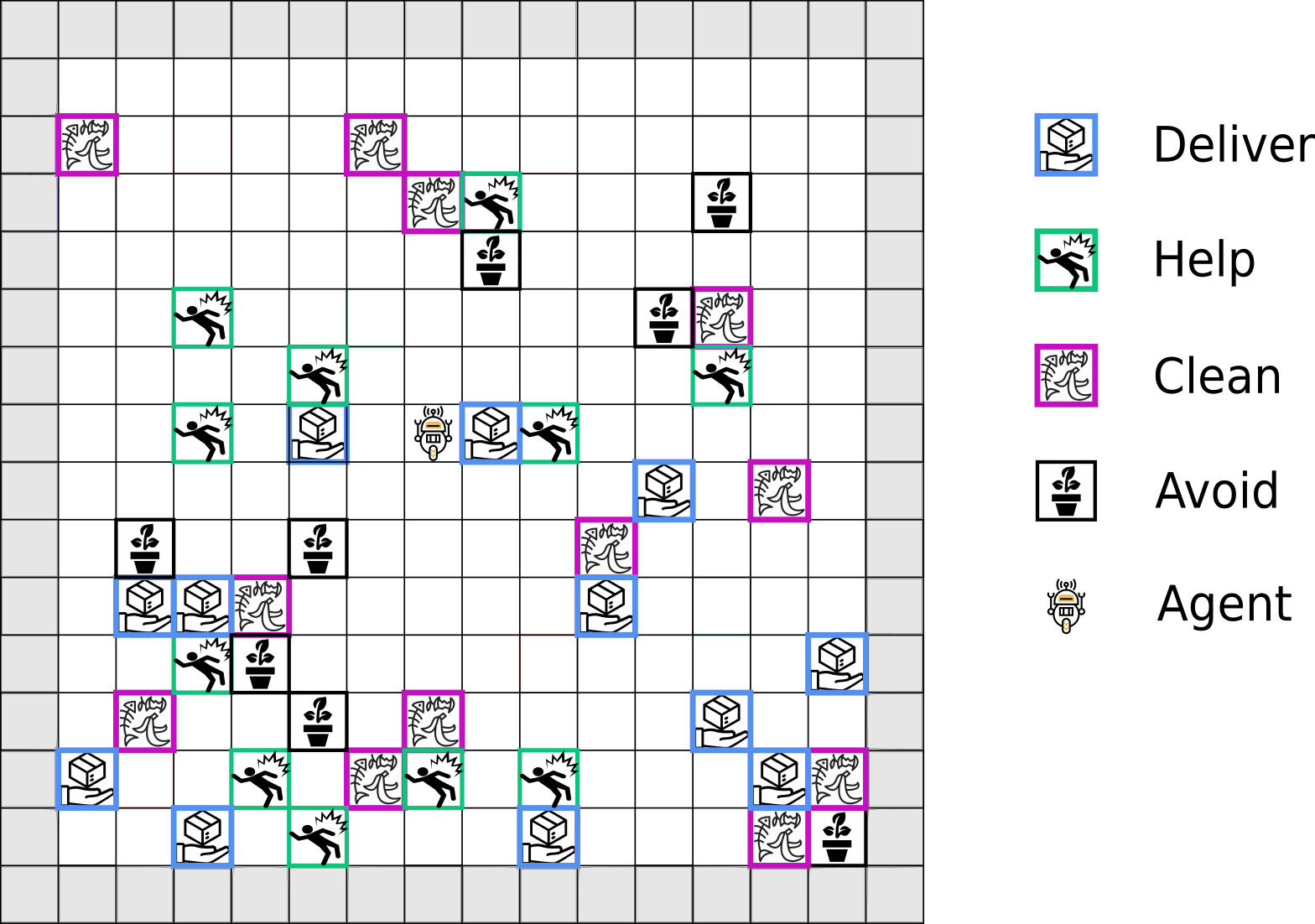}
    \caption{The \textit{Delivery} Environment consists of a primary goal (\textit{Deliver}) and three different norms (\textit{Help} a human, \textit{Clean} a tile, \textit{Avoid} the vase).}
    \label{fig:v3_env}
\end{figure}

\begin{figure*}[t!]
    \centering
    \includegraphics[width=0.99\textwidth]{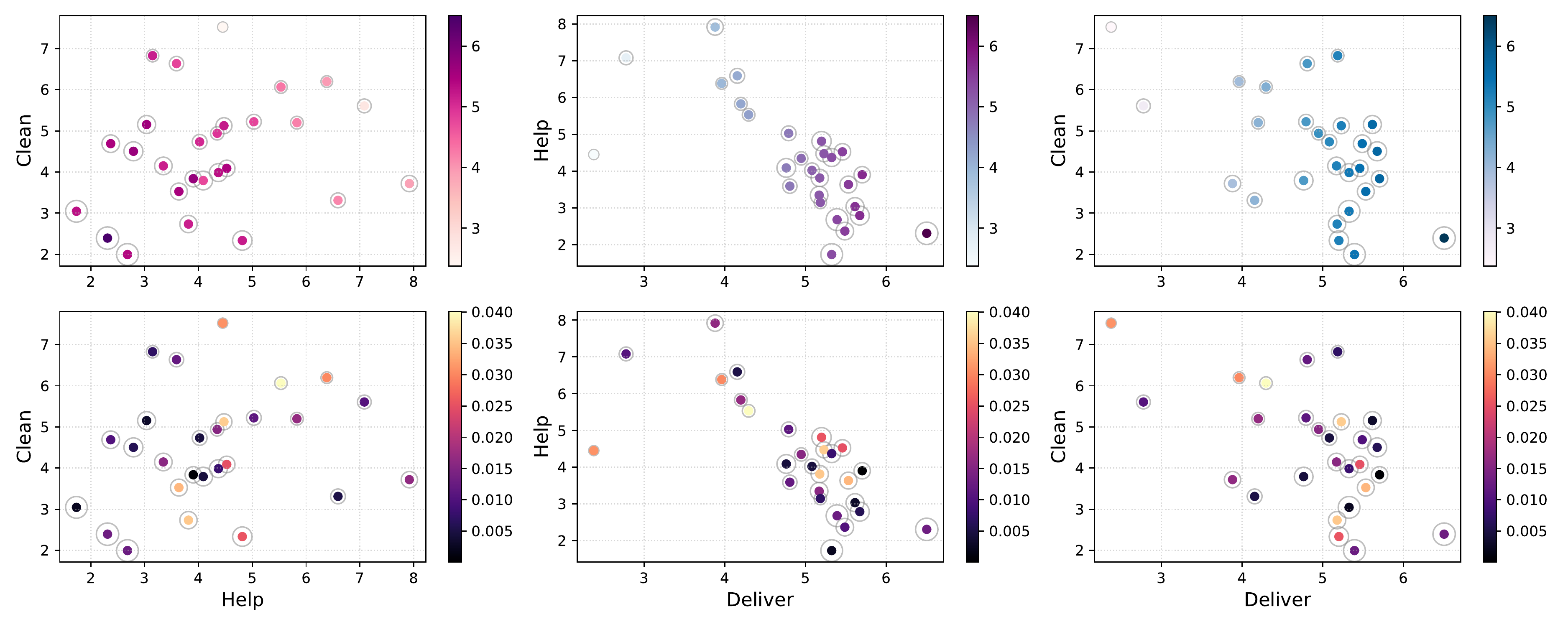}
    \caption{The convex coverage set found by MORAL for three reward dimensions. We plot two-dimensional projections of the attained explicit objectives, with colors indicating the third objective (\textit{top three panels}). The colors in the bottom three panels show the deviation (\ref{eq:kl_prefs}) to the respective preference vector $m$ used during training. Gray circles around each policy indicate the relative amount of broken vases.}
    \label{fig:v3_ccs_123_colors}
\end{figure*}

While the \textit{Emergency} domain illustrated the effectiveness of MORAL in a simplified setting, we yet have to analyze how MORAL performs in larger environments, as well as regarding increased diversity of norms and goals we would like an agent to learn.
To better evaluate MORAL, we therefore define the \textit{Delivery} environment, a randomly initialized $16\times 16$ grid world (figure \ref{fig:v3_env}). As before, the agent has access to the moving, interaction and null actions, but can now encounter a variety of objects. Its primary goal consists of delivering packages to $12$ locations, which is available to the agent via a reward $r_P$ of $+1$ whenever it interacts with a delivery cell. However, there also exist a multitude of people in need of support that the agent can \textit{help} and ``dirty'' tiles that the agent can \textit{clean}. The agent chooses to do so by interacting with each of the respective cells, after which they turn empty. Finally, we randomly place vases throughout the grid, which break whenever the agent steps on their position and can not be interacted with.

Overall, we limit the episode length to $T=50$ time steps and place $12$ of the \textit{help, clean} objectives as well as $8$ vases on the grid. We view this environment as a multi-objective problem including three norms, where \textit{help} and  \textit{clean} are active behaviors, but the ability to \textit{avoid} vases is passive i.e., an inaction. Besides forcing the agent to make trade-offs, this choice allows us to effectively study the quality of solutions found through MORAL, by introducing a symmetry with regard to the three explicit objectives. We assume that preferences are automatically provided by a subjective distribution $m \in \Delta_{\{1,2,3\}}$ encoding desired priorities for \textit{deliver}, \textit{help} and \textit{clean} respectively. Given a pair of trajectories $(\tau_1, \tau_2)$, we then calculate two vectors $s_i = (o_1^i, \dots, o_n^i)$, where $o_k^i$ denotes the obtained returns in terms of the $k$-th objective in trajectory $i$. For example, if $\tau_1$ delivers $3$ packages, helps $1$ person and cleans up $3$ cells, then $s_1 = (3,1,3)$. When normalizing the observed returns into a discrete distribution $\overline{s}_i = s_i/||s_i||_1$, we can provide preferences according to a KL divergence metric
\begin{equation}
    i^* = \argmin_{i\in \{1,2\}} D_{KL}(\overline{s}_i || m). \label{eq:kl_prefs}
\end{equation}
Aside from providing preferences in a principled way, we use this divergence measure to evaluate the overlap between a policy and the provided preferences throughout training. 

We test MORAL using two conflicting demonstration data sets generated by a PPO agent optimizing for (i) helping people and (ii) cleaning tiles, while both try to avoid stepping on vases. As before, we subsequently use MORAL as a regularizer and form $\mathbf{r} = (r_P, f_{\theta_1}, f_{\theta_2})$, where $\theta_1$ and $\theta_2$ denote the trained AIRL parameters. As opposed to the experiment in the \textit{Emergency} domain, there now exists an inherent normative conflict in the demonstrations. Thus, instead of tuning the agent to respect a specific policy that incorporates the normative component into the primary goal, we aim to test whether MORAL is able to retrieve solutions that match a variety of preferences. To achieve this, we vary the supplied preference vector $m$ to match all possible ratios in $\{1,2,3\}^3$ during the active learning stage. Furthermore, we choose to use $25$ queries overall, spread evenly throughout $8\cdot 10^6$ environment steps.

Figure \ref{fig:v3_ccs_123_colors} illustrates the found set of policies, where each point represents a separate run of active learning on different preferences. Since the objective space is three-dimensional, we only show two-dimensional projections and add the third objective through color (figure \ref{fig:v3_ccs_123_colors}, top three panels). Besides this, the number of broken vases is shown by gray circles around each point, where a bigger radius indicates policies that break more vases and a radius of $0$ indicates that no vases are broken on average. To test whether the found policies match the given preferences, we evaluate the KL divergence (\ref{eq:kl_prefs}) of average returns $s$ (generated by the learned policy) to the preference $m$ that was used during training (figure \ref{fig:v3_ccs_123_colors}, bottom three panels). We found that MORAL is overall able to retrieve a diverse set of policies, which accurately represent the different preferences: Firstly, the top three panels show mostly non-dominated policies that span a wide variety of trade-offs, which suggests that MORAL recovers a large part of the convex coverage set. Secondly, the bottom three panels indicate that most of the points achieve a near zero divergence. This means that the agent accurately matches the supplied ratios over objectives. As expected, we also see that the number of broken vases correlates with the weight put on the primary task, since the manually engineered delivery reward is entirely agnostic regarding the vase object. Nonetheless, for appropriate choices of $m$, there exist policies which successfully avoid vases despite delivering an adequate number of packages. These results indicate that when choosing scalarization weights appropriately, minimizing a weighted sum of KL divergences to the respective maximum entropy IRL distributions can achieve implicit normative behaviors without the need of an explicit feedback signal.

\begin{figure}
    \centering
    \includegraphics[width=0.85\columnwidth]{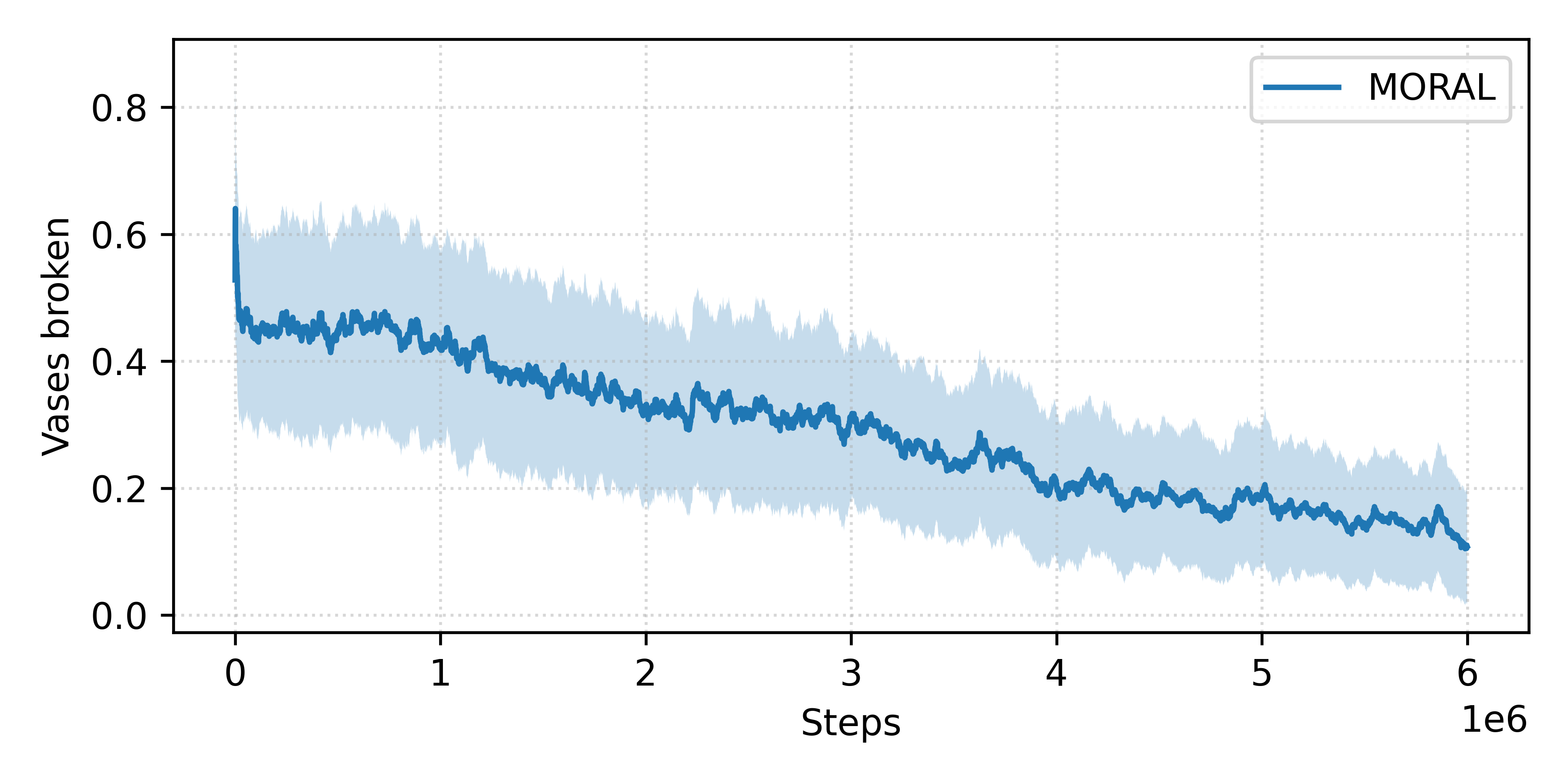}
    \caption{Average number of broken vases over three training runs. MORAL learns a safe policy, despite being provided with adversarial preferences.}
    \label{fig:mal_prefs}
\end{figure}

To investigate the robustness of MORAL against adversarial preferences, we also trained MORAL on $\mathbf{r} = (f_{\theta_1}, f_{\theta_2})$ by giving $25$ preferences such that $\tau_i \succ \tau_j$, whenever $\tau_i$ manages to break more vases. We observe that despite this, the number of broken vases in fact decreases as a function of training steps. Since both experts agree on keeping vases intact, the aggregate reward function cannot be fine-tuned to exhibit the opposite behavior (figure~\ref{fig:mal_prefs}). However, such a guarantee against adversarial preferences only holds when all marginal reward functions induce safe behavior. Under this assumption, this result provides evidence that automatically adhering to common implicit behaviors ensures safety against adversarial preference givers.


\subsection{Comparison to Deep Reinforcement Learning from Human Preferences}

\begin{table}[b!]
\begin{center}
\begin{tabular}{|c|c|c|c|c|} 
 \hline
       & \multicolumn{1}{p{1cm}|}{\centering People Saved} & \multicolumn{1}{p{1.75cm}|}{\centering Extinguished \newline Fire} & \multicolumn{1}{p{1cm}|}{\centering Nr. of \newline Queries} & Steps (IRL) \\ 
       \hline
 MORAL & ${5.76 (\pm 0.13)}$ & $\mathbf{40.08 (\pm 2.9)}$ & 25 & 3e6 (3e6) \\ 
 DRLHP & $5.62 (\pm 0.17)$ & $12.32 (\pm 3.0)$  &$1000$ & 12e6 (-)\\ 
 \hline
 
\end{tabular}
\end{center}
\caption{Comparison of MORAL and DRLHP in \textit{Emergency}.}
\label{tab:pbrlcompv2}
\end{table}

Through its two-step procedure, MORAL is able to combine multiple reward functions from diverse expert behaviors. However, in the active learning stage, we require a single expert to determine which Pareto-optimal policy should ultimately be optimized for. Given enough pairwise comparisons, this directly approximates a policy that best matches the preferences (figure~\ref{fig:moral_path}). For these reasons, among the related RL algorithms, MORAL is most directly comparable to deep reinforcement learning from human preferences (DRLHP) \cite{Christiano2017}, which directly trains a deep reward model from pairwise preferences. To compare the two, we train DRLHP until convergence in \textit{Emergency} and \textit{Delivery} by providing a sufficient number of pairwise comparisons to make up for the missing primary reward and demonstrations that MORAL has access to.

Table \ref{tab:pbrlcompv2} shows the results when providing DRLHP with $1000$ preferences in the same way as MORAL for the \textit{Emergency} domain. As before, trajectories with more people saved are preferred unless equal, in which case extinguishing fire becomes a priority. Although this leads DRLHP to learn a policy that consistently saves most people, it significantly lacks in terms of extinguished fire. This is unsurprising, since DRLHP is not designed to handle multi-objective problems and can not utilize the manually engineered reward signal in any meaningful way. This is because the deep reward model is nonstationary, which poses the combination with the stationary reward $r_P$ to be challenging. As a result, DRLHP needs to maintain a single model for all competing objectives, which can lead to catastrophic forgetting of extinguishing fire when updating the reward network to save more people.

\begin{figure}[t!]
    \centering
    \includegraphics[width=0.99\columnwidth]{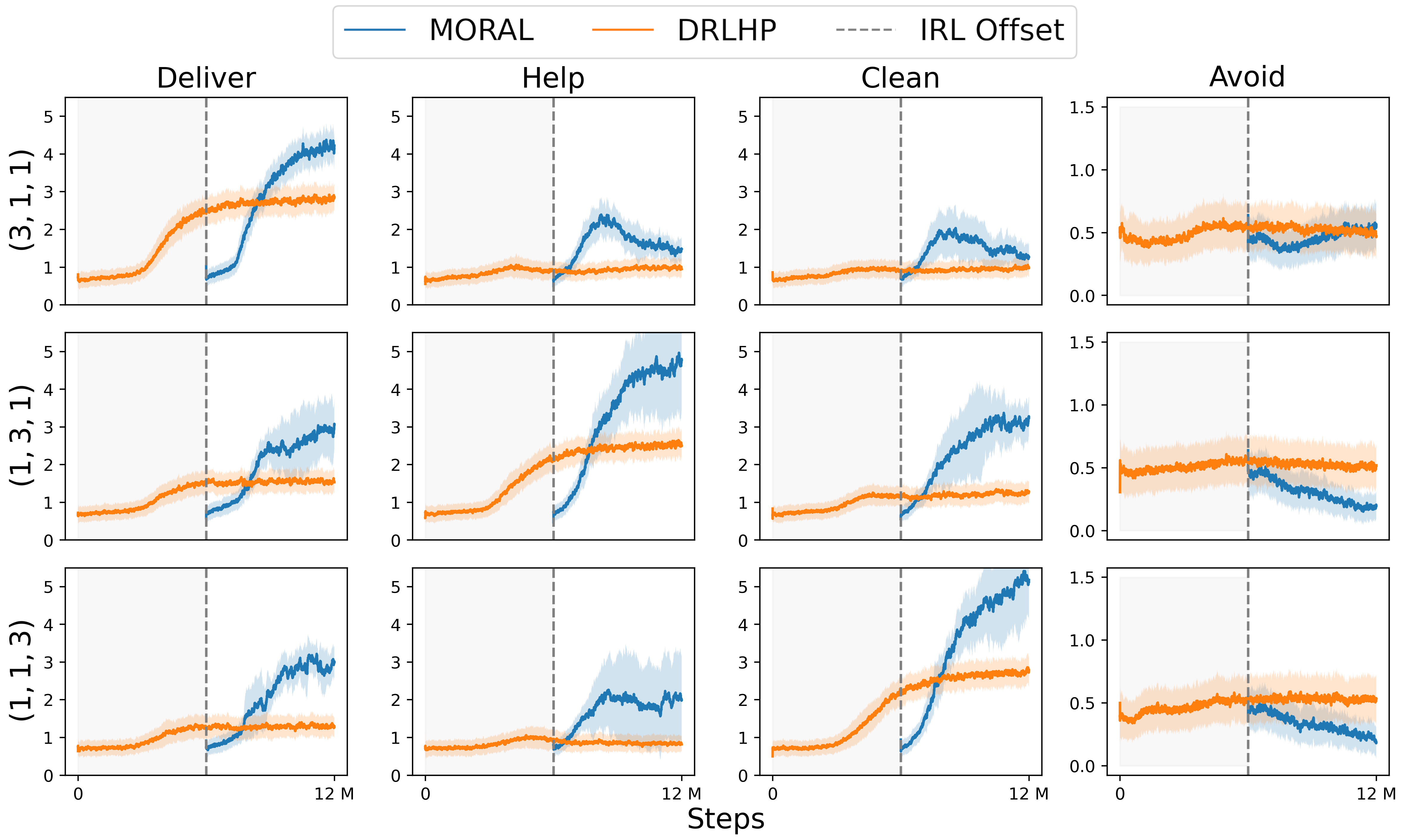}
    \caption{Mean training curves of DRLHP and MORAL on preference ratios $(3,1,1)$ (top), $(1,3,1)$ (middle) and $(1,1,3)$ (bottom).}
    \label{fig:drlhp_comp}
\end{figure}

A similar trend can be observed in the \textit{Delivery} environment, where we compare mean performance of DRLHP versus MORAL on three preference configurations, each of which prefers one of the objectives most strongly. However, since we assumed that avoidance of vases is encoded in the preferences only implicitly, we cannot supply DRLHP with the same set of feedback. Instead, we train DRLHP to prefer trajectories that have a lower mean squared error to the vector of expected returns achieved by MORAL. Figure \ref{fig:drlhp_comp} shows training curves of both methods, where each row represents a preference ratio of $(3,1,1)$, $(1,3,1)$ and $(1,1,3)$ respectively. Aiming to make the comparison fairer, we offset MORAL by the number of total training steps needed for IRL. As before, DRLHP manages to retrieve solutions that loosely resemble the supplied preferences, but fails to converge to Pareto-optimal policies. Furthermore, we notice that for the latter two preferences, sparse objectives such as minimizing the number of broken vases are not picked up by DRLHP. We suspect this to be an exploration issue, where trajectories that break fewer vases are unlikely to arise in queries. Thus, DRLHP optimizes for the remaining objectives as they lead to a higher increase in correctly predicting an expert's preferences. Overall, we conclude that MORAL is more suitable than DRLHP in multi-objective settings that require trading off conflicting objectives from expert data. Nonetheless, MORAL has a theoretical advantage in this environment, since it allows for incorporation of prior knowledge as well as conflicting expert demonstrations. 

\label{sec:drlhp}

\subsection{Ablation}
\begin{figure}[t!]
    \centering
    \includegraphics[width=0.75\columnwidth]{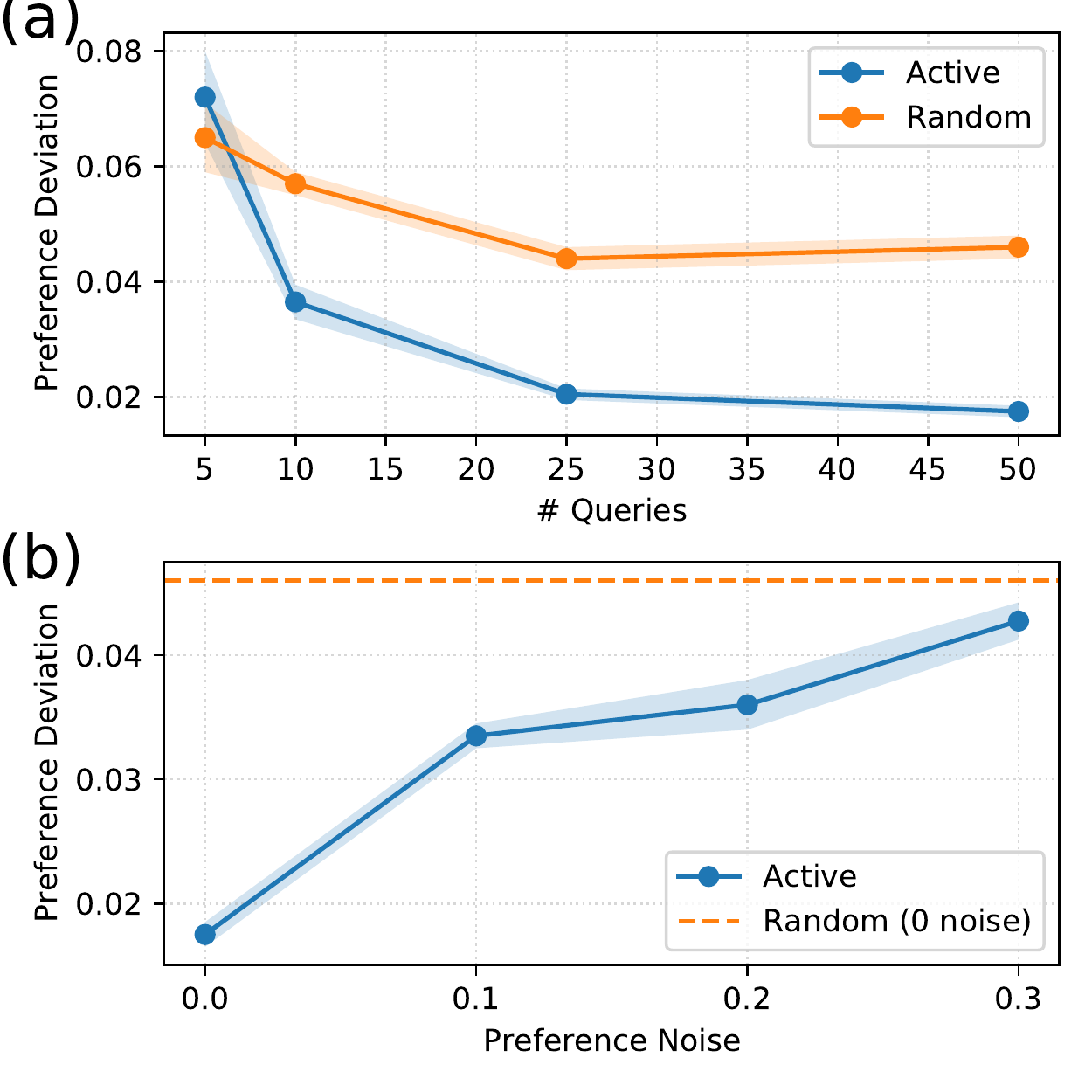}
    \caption{(a) Average preference deviation as a function of the number of active and random queries. (b) Average preference deviation of active learning as a function of the proportion of noisy responses to queries.}
    \label{fig:aamas_ablation}
\end{figure}

In this section, we evaluate MORAL with respect to the necessity of active queries, as well as its robustness against noisy preferences.
To do so, we contrast active versus randomly chosen queries for the same set of preferences as in figure \ref{fig:v3_ccs_123_colors} and plot the average preference deviation (\ref{eq:kl_prefs}) in figure \ref{fig:aamas_ablation} (a). In the case of actively generated queries, there is a clear decrease in preference deviation as a function of total queries. Random queries do not benefit from the increase in queries as much as MORAL, with $50$ queries in fact scoring worse than $25$. We conjecture that this is due to the on-policy sampling of trajectories, which strongly restricts the space of trajectories that the agent can query for. On the other hand, MORAL will, always seek pairs of trajectories that naturally exhibit larger variances between competing goals in order to generate queries with high information content. When ensuring that the agent maintains adequate levels of exploration throughout training, this suffices to ensure a decrease in deviation even in the large query regime.



To investigate the robustness of MORAL in the presence of contradictory feedback, we train policies on the same set of preferences as before in figure \ref{fig:aamas_ablation} (b), but provide random answers to each of the $50$ queries with a certain probability. Unsurprisingly, we see a sharp increase in deviation when injecting a noise level of $0.1$, above which the growth in error diminishes. Nonetheless, active queries with a random answer probability of $0.3$ still retrieve slightly more accurate representations than random queries without any noise. Such robustness with respect to noise is important, since our experiments only cover synthetic simulation studies, whereas human feedback is unlikely to be as consistent. Even though random noise is only an approximation of human error, we conclude from our results that seeking volume removal in the active learning loop does not make the algorithm more susceptible to converging to locally optimal scalarization weights in this case.

\section{Related Work}
\noindent \textbf{Machine Ethics} Using the notion of uncertainty and partial observability, RL has been suggested as a framework for ethical decision-making \cite{Abel2016ReinforcementLA}. We frame the problem of learning norms in a multi-objective context, which can be interpreted as inducing partial observability over the set of reward scalarizations one would wish to optimize. Overall, the motivation behind our approach is conceptually similar to policy orchestration \cite{noothigattu2019teaching} and ethics shaping \cite{Wu2018} (table~\ref{tab:MORAL_COMP}). Policy orchestration \cite{noothigattu2019teaching} also adopts a multi-objective view to incorporate ethical values into reward-driven RL agents, by solving a bandit problem that uses IRL to alternately play ethical and reward maximizing actions. Similarly to policy orchestration, MORAL employs a two-step procedure. However, besides the use of deep RL, MORAL differs since it learns to combine reward functions, whereas policy orchestration learns to combine policies. This allows MORAL to learn Pareto optimal policies at the cost of interpretability.
Furthermore, policy orchestration requires a manual specification of the scalarization parameter, which MORAL can automatically infer through the use of active learning.

Ethics shaping \cite{Wu2018} learns a reward shaping term from demonstrations, but does not scale beyond manually engineered features. Besides that, their approach is constrained to learning from a single expert. Although AIRL has been previously suggested to alleviate the issue of scalability \cite{Peschl2021}, a method that is able to trade off rewards from multiple sources has not yet been developed in this setting. Finally, Ecoffet et al. \cite{Ecoffet2020} suggest a sequential voting scheme for learning conflicting values, but it requires explicit encoding of the different values at stake.

\vspace{4pt}

\textbf{Inverse Reinforcement Learning} Similarly to related IRL research, our work builds on AIRL \cite{Fu2017} for inferring rewards from a multimodal distribution of demonstrations. Unlike previous research, which has focused on introducing latent variable models \cite{li2017infogail, hausman2017multi, yu2019meta, gruver2020multi, venuto2020oirl, sharma2018directed} in the context of multiagent, multitask and hierarchical reward learning, we instead focus on the combination of labeled demonstration data. As such, our setup is similar to Gleave and Habryka \cite{gleave2018multi} and Xu et al. \cite{Xu2019LearningAP}, where a reward function is meta-learned by having explicit access to different task distributions. However, we learn from demonstrations in a multi-objective context, which has, to our knowledge, not yet been studied before. 

Besides this, IRL has been applied in the context of value alignment \cite{Hadfield-Menell2016}, where inverse reward design (IRD) \cite{Hadfield-Menell2017} has been proposed to learn a distribution of reward functions through IRL that leverages uncertainty to avoid unintended behavior. Although we also learn a distribution over reward functions, IRD focuses on finding safe goal specifications from a single reward function, whereas we study the extraction of value-aligned policies from a multitude of demonstrations. As a result, our research is more similar to multitask IRD \cite{krasheninnikov2021combining}, which studies formal criteria for combining reward functions from multiple sources. We, on the other hand, drop the formal assumptions and propose a practical method for combining reward functions learned through deep neural networks.

\vspace{4pt}

\noindent \textbf{Learning from Expert Feedback} Besides IRL, there exist a variety of approaches for training RL agents from expert data, including scalar-valued input \cite{knox2009interactively,warnell2018deep}, natural language \cite{AlNahian2020}, intervention \cite{Saunders2018} and pairwise preferences \cite{Wirth2017, Christiano2017}. Similarly to Christiano et al. \cite{Christiano2017}, we employ a Bradley-Terry model for training a nonstationary reward function by comparing trajectories from on-policy RL experience. However, our model of pairwise preferences operates on a set of abstract high-level reward functions, whereas \cite{Christiano2017} learn a single end-to-end reward model. Furthermore, our approach combines demonstration and preference data, which is more similar to Ibarz et al. \cite{Ibarz2018RewardLF}. Nonetheless, \cite{Ibarz2018RewardLF} uses demonstration data for pretraining a preference-based reward model, which does not account for conflicting demonstrations. MORAL, on the other hand, allows for the inclusion of multiple experts as well as prior knowledge, thus making it suitable for resolving normative conflicts. 

Preference-based reward learning with multiple experts has been recently studied by Myers et al. \cite{myers2021learning}. They propose an active learning algorithm for efficiently learning multimodal reward functions that can represent different preferences. This differs from our work in two key points: Firstly, aside from the incorporation of demonstrations, we show how to use active learning to find a trade-off between conflicting reward functions. Secondly, unlike their work, we study how to learn a policy alongside the reward function.

Finally, by combining different expert reward functions, we have shown that MORAL interpolates between maximum-entropy IRL distributions. From this point of view, we consider our work a counterpart to Brown et al. \cite{brown2019extrapolating}, which ranks demonstration data in order to extrapolate beyond the behavior of a single expert.

\vspace{4pt}

\begin{table}[b!]
    \centering
    \begin{tabular}{|c|c|c|c|c|}
    \hline
       & \multicolumn{1}{p{1cm}|}{\centering Ethics- \newline Shaping \cite{Wu2018}}  & \multicolumn{1}{p{1.7cm}|}{\centering Policy- \newline Orchestration \cite{noothigattu2019teaching}} & \multicolumn{1}{p{1cm}|}{\centering DRLHP \newline \cite{Christiano2017}}  & MORAL\\
        \hline
         Deep & \multirow{2}*{$\Large\color{red}{\pmb\times}$}  & \multirow{2}*{$\Large\color{red}{\pmb\times}$} & \multirow{2}*{$\Large\color{green}{\checkmark}$} & \multirow{2}*{$\Large\color{green}{\checkmark}$} \\
         Learning & & & & \\
         \hline 
         Multi- &  \multirow{2}*{$\Large\color{blue}{\sim}$}  & \multirow{2}*{$\Large\color{green}{\checkmark}$}&  \multirow{2}*{$\Large\color{red}{\pmb\times}$} &  \multirow{2}*{$\Large\color{green}{\checkmark}$} \\
         Objective & & & &\\
         \hline
         Multiple  &  \multirow{2}*{$\Large\color{red}{\pmb\times}$}  &  \multirow{2}*{$\Large\color{blue}{\sim}$} &  \multirow{2}*{$\Large\color{red}{\pmb\times}$} &  \multirow{2}*{$\Large\color{green}{\checkmark}$}\\
         Experts & & & &\\
         \hline
    \end{tabular}
    \caption{Comparison of MORAL to previous work in terms of supported capabilities.}
    \label{tab:MORAL_COMP}
\end{table}

\noindent \textbf{Multi-Objective Decision-Making}
Typically, MORL algorithms trade off multiple objectives by learning a policy, or a set thereof, that can represent a range of Pareto-optimal solutions \cite{roijers2013, Yang2019}. On the other hand, our model learns a distribution over reward functions, which interactively guides the search to produce a single Pareto-optimal policy. Aside from sample efficiency, this mitigates the problem of varying reward scale, which has previously been addressed by multi-objective maximum a posteriori policy optimization (MO-MPO) \cite{abdolmaleki2020distributional}. However, MO-MPO requires explicit preferences over objectives, which is not always feasible when combining learned rewards that are inherently difficult to compare. 

By using on-policy RL experience to learn scalarization weights, MORAL can be viewed as an interactive MORL algorithm. To date, interactive MORL has mainly been applied to bandits for linear \cite{Roijers2017InteractiveTS} and nonlinear \cite{Roijers2020InteractiveMR} transformations of the reward components, but has not yet been studied in the full RL setting. We believe that this is the case because MORL research usually assumes environments with manually engineered reward functions, in which big parts of the Pareto boundary exhibit interesting solutions. In the case of trading off learned reward functions, however, we suggest that our interactive approach poses a more adequate option.

\section{Discussion}
In our work, we propose MORAL, a method for combining learned reward functions from multiple experts. We have shown MORAL to be a technical approach for aligning deep RL agents with human norms, which uses active learning to resolve value conflicts within expert demonstrations. We consider our research a step towards multi-objective RL with learned rewards, which has not yet been addressed before. Previous approaches such as ethics-shaping \cite{Wu2018} and policy orchestration \cite{noothigattu2019teaching} have highlighted the strength of combining reward functions with expert demonstrations for value alignment, whereas DRLHP \cite{Christiano2017} has demonstrated the scalability of deep preference-based RL (table \ref{tab:MORAL_COMP}). MORAL unifies these ideas into a single method, which allows it to be applied in the presence of deep function approximation and multiple experts. This theoretical advantage is reflected in our experiments, which show that, unlike DRLHP, MORAL succeeds in retrieving Pareto-optimal solutions. Furthermore, we have shown MORAL to automatically learn implicit social norms if expert demonstrations agree on them. This informativeness about desirable behavior has been previously identified as a desideratum for combining reward information by Krasheninnikov et al. \cite{krasheninnikov2021combining} and we have shown that it allows the active queries to focus on higher-level normative conflicts.

Nonetheless, several avenues for future research remain to be addressed. Firstly, generating queries from on-policy experience puts MORAL at risk of local optimaility. In sparse environments, we therefore consider the introduction of a separate exploration policy for active learning to be useful. Secondly, combining multiple forms of expert supervision is challenging, due to a risk of accumulating errors and modelling assumptions for each type of input. We suppose further research in AIRL will be necessary to prevent overfitting of the reward network. Similarly to Gleave and Habryka \cite{gleave2018multi}, we found the reoptimization of AIRL reward functions to decrease performance, indicating that the learned rewards are entangled with the state distribution of the generator policy. Although this will require significant progress in deep IRL, we expect future methods to be easily integrated into MORAL by replacing AIRL. Furthermore, one could pursue unsupervised techniques to extend MORAL to unlabeled demonstration datasets. When learning social norms from large scale real-world demonstration data, it might be infeasible to learn separate reward functions for each expert. Unsupervised learning of reward functions that correspond to the different modes of behavior instead could alleviate this issue.

Overall, our research highlights the importance of multi-objective sequential decision-making without explicitly provided reward functions. Aside from value alignment \cite{vamplew2018human}, the ability to detect and respond to a divergence in values has been recognized as a central trait for building human-like AI \cite{booch2020thinking}. Further, following the principle of meaningful human control \cite{deSio2018}, MORAL can contribute to increase an agent's responsiveness to conflicting human norms, while maintaining human autonomy in determining desired trade-offs. This research contributes to the broader goal of designing and developing safe AI systems that can align to human values and norms.

\bibliographystyle{ACM-Reference-Format} 
\bibliography{sample}


\newpage
\appendix 

\section{Architectures}
This section describes all network architectures used throughout the experiments. For ease of exposition, we omit grid world dimensionalities of each environment, and instead only report the amount of output channels and kernel sizes respectively. Furthermore, each convolutional layer uses a stride of $1$ and no padding, which we found sufficient due to the relatively small sizes of the grids.

\subsection{Proximal Policy Optimization}
\label{sec:PPOArch}
\begin{figure}[h!]
    \centering
    \includegraphics[width=0.99\columnwidth]{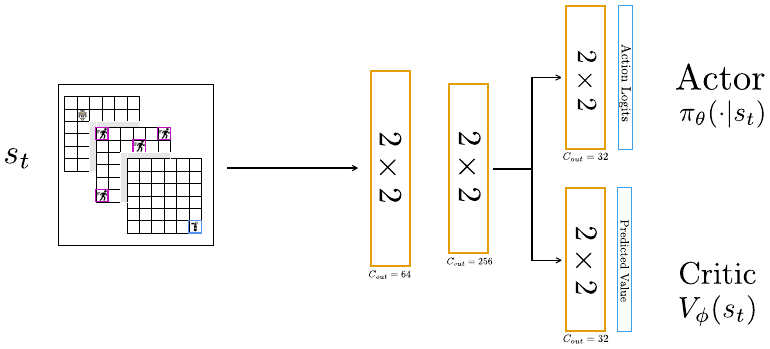}
    \caption{Actor-Critic architecture of the PPO agent consisting of convolutional (yellow) and linear (blue) layers. Regardless of the input dimension, we use $C_{out}$ output channels and kernel sizes of $2$.}
    \label{fig:ppoarch}
\end{figure}

For PPO, we employ a convolutional actor-critic architecture with shared base layers, as illustrated in figure \ref{fig:ppoarch}. We use two convolutional layers to form a feature array with $256$ channels, which is then passed to the actor and critic in parallel. Finally, the actor employs a linear layer with output dimensions equal to the number of actions $|\mathcal{A}| =9$ on the flattened feature representations of the final convolutional layer. Similarly, the critic employs a final linear layer with a scalar output to predict the current value. To draw action samples from the actor, a softmax is performed over its last linear layer, and we treat the resulting vector as a categorical distribution. In between layers, we employ standard ReLU activations to facilitate nonlinearity.

\subsection{Reward Network - Emergency}
\begin{figure}[h!]
    \centering
    \includegraphics[width=0.99\columnwidth]{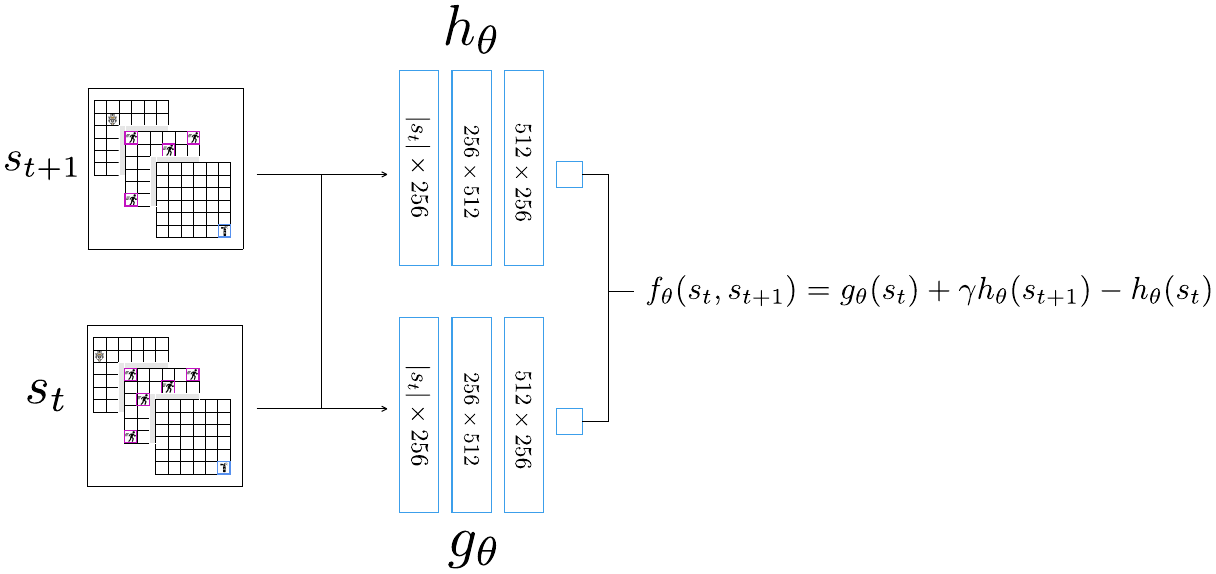}
    \caption{Linear discriminator architecture. A forward pass calculates activations of the networks $g_\theta$ and $h_\theta$ respectively and combines them into the reward prediction $f_\theta$.}
    \label{fig:mlpdisc}
\end{figure}

Due to the small size of the environment, we employ a dense neural network for the AIRL discriminator architecture in \textit{Emergency}. In this case, we flatten grid world states into a single vector and pass them through each layer, as illustrated in figure \ref{fig:mlpdisc}. Similarly to \cite{Fu2017}, we decompose the network output $f_\theta(s, s') = g_\theta(s) + \gamma h_\theta(s') - h_\theta(s)$, where $\gamma \in [0,1]$ is the discount factor. While \cite{Fu2017} propose this decomposition to retrieve a state-only reward function, we instead only use it for matching the AIRL implementation, but use $f_\theta$ as the reward function in subsequent steps.

\subsection{Convolutional Reward Network - Delivery}
\label{sec:ConvRewArch}
\begin{figure}[h!]
    \centering
    \includegraphics[width=0.99\columnwidth]{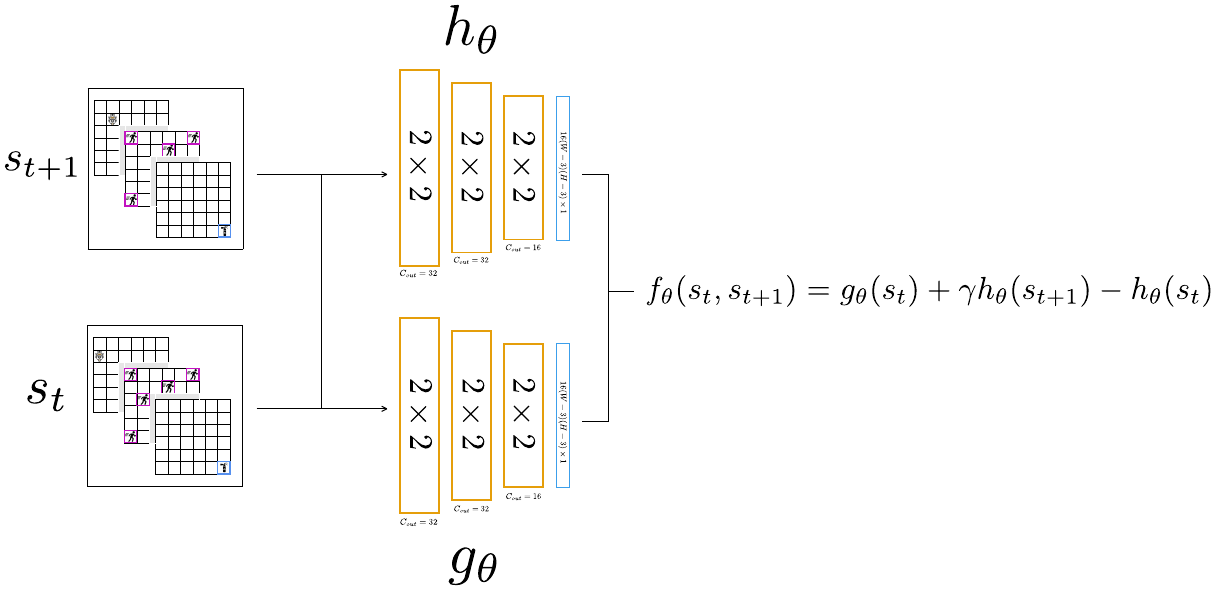}
    \caption{Convolutional discriminator architecture for training AIRL in bigger environments with a parallel stream of convolutional (yellow) and linear (blue) layers.}
    \label{fig:convdiscarch}
\end{figure}

We found the MLP architecture of the AIRL discriminator shown in figure \ref{fig:mlpdisc} to be insufficient in larger grid world sizes. For this reason, we employ a convolutional reward network in \textit{Delivery} as shown in figure \ref{fig:convdiscarch}. In principle, the network follows the same structure as in \textit{Emergency}, but replaces linear layers with convolutional ones. To do so, we employ three convolutional layers followed by a single linear layer that acts on respective flattened feature maps for both $h_\theta$ and $g_\theta$ and form our reward estimate as before. Finally, we use LeakyReLU activations with a slope of $\alpha=0.01$ on all hidden layers.

\subsection{Deep Reinforcement Learning from Human Preferences}
\label{sec:appendix_DRLHP_arch}
\begin{figure}[h!]
    \centering
    \includegraphics[width=0.99\columnwidth]{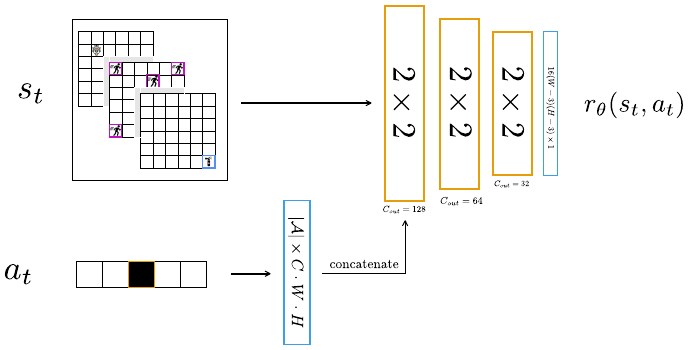}
    \caption{Reward model architecture for DRLHP with convolutional (yellow) and linear (blue) layers. Actions are embedded through a linear layer and concatenated with the current state before being fed through subsequent layers.}
    \label{fig:drlhparch}
\end{figure}

For DRLHP we train a PPO agent using the architecture shown in figure \ref{fig:ppoarch} in parallel with a deep reward model, which we show in figure \ref{fig:drlhparch}. The reward model takes a state-action pair at each time step and outputs a predicted reward $r_\theta(s_t,a_t)$. We first one-hot encode the action $a_t$ and then embed it into a vector with the same dimensionality as the input state $s_t$. To do so, we train a linear embedding layer with output dimensions $C\cdot W\cdot H$, where $(C,W,H)$ denote the amount of channels, width and height of the state $s_t$ respectively. Embedded actions then get reshaped and concatenated with $s_t$ along the channel dimension to form an array of dimension $(2C, W, H)$ (batch dimension omitted). This array is fed through three convolutional layers with $128$, $64$ and $32$ output channels respectively. Finally, the resulting flattened feature maps are processed by a linear layer to produce the reward estimate. As in the AIRL discriminator architecture, we employ LeakyReLU activations with a slope parameter $\alpha=0.01$.

\section{Proof of Theorem 2.1}
By definition of the Kullback-Leibler divergence, we have  
\begin{equation}
    D_{KL}(\pi(\tau)||p_{\theta_i}(\tau))
    = \mathbb{E}_\pi\left[\sum_{t=0}^T   \log \pi(a_t|s_t) -r_{\theta_i}(s_t, a_t) \right] + \log Z_{\theta_i},
\end{equation}
where $Z_\theta = \int \overline{p}_\theta(\tau)d\tau$ is the partition function.
Using $\sum\mathbf{w}_i = 1$, we can now take the weighted sum of Kullback-Leibler divergences and obtain

\begin{align*} 
&\sum_{i=1}^k w_i \mathbb{E}_\pi \left[\sum_{t=0}^T \log \pi(a_t | s_t) - r_{\theta_i}(s_t,a_t) \right] +  \sum_{i=1}^k w_i\log Z_{\theta_i} \\
    &= \mathbb{E}_\pi \left[\sum_{t=0}^T \log \pi (a_t|s_t) - \left(\sum_{i=1}^k w_i r_{\theta_i}(s_t,a_t)\right)  \right]  + \sum_{i=1}^k w_i\log Z_{\theta_i}.
\end{align*}
\noindent Minimizing over $\pi$ yields the desired expression, since the normalization functions $Z_{\theta_i}$ as well as the weights $w_i$ are constants as a function of the policy $\pi$. $\hfill \blacksquare$

\section{Markov Chain Monte Carlo}
In our implementation, we follow the procedure by \cite{Sadigh2017ActivePL} for simplifying the approximation of the Bradley-Terry posterior using Markov Chain Monte Carlo (MCMC). Namely, instead of using the original likelihood

\begin{equation}
    p(\tau_i \succ \tau_j|\mathbf{w}) = \frac{\exp({\mathbf{w}^T \mathbf{r}(\tau_i)})}{\exp({\mathbf{w}^T \mathbf{r}(\tau_j)}) + \exp({\mathbf{w}^T \mathbf{r}(\tau_i)})},
\end{equation}
we instead opt for a proxy likelihood of the form

\begin{equation}
    \hat{p}(\tau_i\succ \tau_j|\mathbf{w}) = \min (1, \exp (\mathbf{w}^T\Delta_{ij})),
    \label{eq:proxy_likelihood}
\end{equation}
where $\Delta_{ij} = \mathbf{r}(\tau_i) - \mathbf{r}(\tau_j)$. Its mode always evaluates to $0$,
which allows us to efficiently obtain posterior estimates through the Metropolis-Hastings algorithm \cite{chib1995understanding} with a warm start to the distribution mode.

\section{Hyperparameters - Emergency}
In the following, we will list all hyperparameter configurations used for the experiments of the main sections. Aside from algorithm specific hyperparameters, we always employ a learning rate for PPO (\textit{lr-PPO}) that determines the gradient step size used in the agent's Adam optimizer, a trust region clip parameter ($\epsilon$-\textit{clip}) that determines how far the updated policy is allowed to diverge from the old, a time discounting parameter $\gamma$, the amount of gradient steps taken on the policy loss per epoch (\textit{Epochs PPO}) and the amount of environment episodes used for each epoch in PPO (\textit{Batch Size PPO}). All policies were trained in a vectorized environment with $12$ instances for \textit{Environment Steps} amount of interactions.
\label{sec:appendix_v2_hyperparam}
\subsection{AIRL}
In AIRL, we use an Adam optimizer with its own learning rate for the discriminator (\textit{lr-Discriminator}). Furthermore, \textit{Batch Size Discriminator} determines the amount of state-action pairs used in a single training batch. Hyperparameters are reported in table \ref{tab:hyperparams_airl_singlenorm}.

\begin{table}[h]
    \centering
    \begin{tabular}{|c|c|}
    \hline
    Hyperparameter & Value \\
    \hline\hline
        lr-Discriminator & 5e-4\\
        lr-PPO & 5e-4\\
        Batch Size Discriminator & 512 \\
        Batch Size PPO & 12\\
        Environment Steps & 3e6\\
        $\epsilon$-clip & 0.1 \\
        $\gamma$ & 0.999\\
        Epochs PPO & 5\\ 
        \hline
    \end{tabular}
    \caption{AIRL hyperparameters in \textit{Emergency}.}
    \label{tab:hyperparams_airl_singlenorm}
\end{table}

\subsection{Active Learning}
In the active learning step of MORAL, we query at fixed time intervals with a prespecified amount of total queries (\# \textit{Queries}) that get evenly distributed across the amount of available environment steps. Besides that, no additional hyperparameters are necessary. However, we note that if the posterior converges to a local optimum prematurely, one can employ a normalization parameter $c>0$ to multiply the vector valued reward function $\overline{\mathbf{r}}(s,a) := c\cdot \mathbf{r}$. For small choices of $c$, one can expect to make the posterior less sensitive to updates at each step. Nonetheless, we found an inclusion of such hyperparameter to be unnecessary in our experiments, since marginal reward functions are normalized by their respective optimal values regardless. We report active learning hyperparameters in table \ref{tab:hyperparams_moral_singlenorm}.

\begin{table}[h!]
    \centering
    \begin{tabular}{|c|c|}
    \hline
    Hyperparameter & Value \\
    \hline\hline
        lr-PPO & 3e-4\\
        \# Queries & 25\\
        Batch Size PPO & 12\\
        Entropy Regularization & 0.25\\
        Environment Steps & 6e6\\
        $\epsilon$-clip & 0.1 \\
        $\gamma$ & 0.999\\
        Epochs PPO & 5\\ 
        \hline
    \end{tabular}
    \caption{Active learning hyperparameters in \textit{Emergency}.}
    \label{tab:hyperparams_moral_singlenorm}
\end{table}

\subsection{DRLHP}
To make DRLHP conceptually similar to MORAL, we employ queries at constant time intervals using a fixed amount of total queries (\# \textit{Queries}) across the available environment steps. Besides that, we update the deep reward model after a constant amount of environment steps (\textit{Update Reward Model Frequency}) with the Adam optimizer and a corresponding learning rate (\textit{lr-Reward Model}). Overall, higher entropy regularization was necessary to ensure adequate exploration for learning an accurate reward model. We report DRLHP hyperparameters in table \ref{tab:hyperparams_pbrl_singlenorm}.

\begin{table}[h!]
    \centering
    \begin{tabular}{|c|c|}
    \hline
    Hyperparameter & Value \\
    \hline\hline
        lr-PPO & 3e-4\\
        lr-Reward Model & 3e-5\\
        Update Reward Model Frequency & 50\\
        \# Queries & 1000\\
        Batch Size PPO & 12\\
        Batch Size Reward Model & 32\\
        Entropy Regularization & 1\\
        Environment Steps & 12e6\\
        $\epsilon$-clip & 0.1 \\
        $\gamma$ & 0.999\\
        Epochs PPO & 5\\ 
        \hline
    \end{tabular}
    \caption{Hyperparameter setup for DRLHP in \textit{Emergency}.}
    \label{tab:hyperparams_pbrl_singlenorm}
\end{table}

\section{Hyperparameters - Delivery}
In \textit{Delivery}, the choice of hyperparameters is similar, besides a consistent increase in environment steps due to a higher task complexity. 
\label{sec:appendix_v3_hyperparam}
\subsection{AIRL}
\label{sec:appendix_v3_airl_hyperparam}
To avoid overfitting and balance the discriminator and generator performances, we lower the learning rate of the discriminator. We show AIRL hyperparameters for \textit{Delivery} in table \ref{tab:hyperparams_airl_v3}.

\begin{table}[h]
    \centering
    \begin{tabular}{|c|c|}
    \hline
    Hyperparameter & Value \\
    \hline\hline
        lr-Discriminator & 5e-5\\
        lr-PPO & 5e-4\\
        Batch Size Discriminator & 512 \\
        Batch Size PPO & 4\\
        Environment Steps & 6e6\\
        $\epsilon$-clip & 0.1 \\
        $\gamma$ & 0.999\\
        Epochs PPO & 5\\ 
        \hline
    \end{tabular}
    \caption{AIRL hyperparameters in \textit{Delivery}.}
    \label{tab:hyperparams_airl_v3}
\end{table}

\subsection{Active Learning}
The following table shows the typical hyperparameter setup for the active learning step of MORAL. Note, however, that while the amount of total environment steps were held fixed throughout different runs, the total number of queries varied, as described in the respective experiments. See table \ref{tab:hyperparams_moral_v3} for the active MORL hyperparameters in \textit{Delivery}.

\begin{table}[h!]
    \centering
    \begin{tabular}{|c|c|}
    \hline
    Hyperparameter & Value \\
    \hline\hline
        lr-PPO & 3e-4\\
        \# Queries & 25\\
        Batch Size PPO & 12\\
        Entropy Regularization & 0.25\\
        Environment Steps & 8e6\\
        $\epsilon$-clip & 0.1 \\
        $\gamma$ & 0.999\\
        Epochs PPO & 5\\ 
        \hline
    \end{tabular}
    \caption{Active learning hyperparameters in \textit{Delivery}.}
    \label{tab:hyperparams_moral_v3}
\end{table}

\subsection{DRLHP}
To ensure that the DRLHP has a comparable amount of available information about the expert's underlying preferences, we provide $5000$ overall queries over the course of training. We show DRLHP hyperparameters for \textit{Delivery} in table \ref{tab:hyperparams_pbrl_v3}.

\begin{table}[h!]
    \centering
    \begin{tabular}{|c|c|}
    \hline
    Hyperparameter & Value \\
    \hline\hline
        lr-PPO & 3e-4\\
        lr-Reward Model & 3e-5\\
        Update Reward Model Frequency & 50\\
        \# Queries & 5000\\
        Batch Size PPO & 12\\
        Batch Size Reward Model & 12\\
        Entropy Regularization & 1\\
        Environment Steps & 12e6\\
        $\epsilon$-clip & 0.1 \\
        $\gamma$ & 0.999\\
        Epochs PPO & 5\\ 
        \hline
    \end{tabular}
    \caption{Hyperparameter setup for DRLHP in \textit{Delivery}.}
    \label{tab:hyperparams_pbrl_v3}
\end{table}

\end{document}